\documentclass{article}
\usepackage{spconf}

\usepackage{appendix}
\usepackage{makecell}

\usepackage[english]{babel}

\usepackage{ifpdf}

\usepackage{cite} 
\usepackage{url}
\usepackage{hyperref}

\usepackage[pdftex]{graphicx}
\graphicspath{{./figures/}}
\usepackage{color}
\usepackage{pgf, tikz, pgfplots}
\usetikzlibrary{shapes, arrows, automata, plotmarks}
\usetikzlibrary{calc,hobby,decorations}
\usepackage{fixltx2e}
\usepackage{stfloats}

\usepackage[cmex10]{amsmath}
\usepackage{amsfonts, amssymb, amsthm}
\usepackage{mathrsfs}

\usepackage{algorithm,algpseudocode}
\algnewcommand{\LeftComment}[1]{\Statex \(\triangleright\) #1}

\usepackage{color, colortbl}
\definecolor{Gray}{gray}{0.9}
\usepackage{array}

\usepackage{enumitem}

\usepackage[margin=16mm]{geometry}
\usepackage{multirow}
\usepackage{subcaption}

\input{pennColors.sty}
\usepackage{needspace}

\input{mySymbol.sty}

\newtheorem{remark}{\hspace{0pt}\bf Remark}

\newcommand{\QED}{\hfill\ensuremath{\blacksquare}}

\pdfoutput=1

\title{GRAPH-TIME CONVOLUTIONAL NEURAL NETWORKS}
\name{Elvin Isufi and Gabriele Mazzola\thanks{The authors are with the Intelligent Systems Department, Delft University of Technology, Delft, The Netherlands. Corresponding author: E. Isufi. e-mails: e.isufi-1@tudelft.nl; gabriele.m1995@gmail.com}
\address{\vspace{-.5cm}}
}
\begin{document}
\ninept
\maketitle
\begin{abstract}
Spatiotemporal data can be represented as a process over a graph, which captures their spatial relationships either explicitly or implicitly. How to leverage such a structure for learning representations is one of the key challenges when working with graphs. In this paper, we represent the spatiotemporal relationships through product graphs and develop a first principle graph-time convolutional neural network (GTCNN). The GTCNN is a compositional architecture with each layer comprising a graph-time convolutional module, a graph-time pooling module, and a nonlinearity. We develop a graph-time convolutional filter by following the shift-and-sum principles of the convolutional operator to learn higher-level features over the product graph. The product graph itself is parametric so that we can learn also the spatiotemporal coupling from data. We develop a zero-pad pooling that preserves the spatial graph (the prior about the data) while reducing the number of active nodes and the parameters. 
Experimental results with synthetic and real data corroborate the different components and compare with baseline and state-of-the-art solutions.
\end{abstract}
\begin{keywords}
Graph signal processing; graph neural networks; graph-time neural networks; spatiotemporal learning.\vspace{-.25cm}
\end{keywords}

\section{Introduction}
\label{sec:intro}\vspace{-.25cm}

Multivariate temporal data provide unique challenges to the learning algorithms because of their intrinsic spatiotemporal dependencies. These dependencies can be captured by a graph either explicitly such as in sensor or social networks or implicitly such as in recommender systems. This graph represents the spatial coupling between data, which translates into a tantamount graph-temporal coupling. The learning algorithm should, therefore, be equipped with effective biases to exploit this structure for learning spatiotemporal representations.
Building on recent advances in processing and learning over graphs \cite{ortega2018graph,hamilton2017representation}, different solutions have been proposed to learn from spatiotemporal data \cite{wang2020deep}. The key to learning is the algorithm's ability to embed spatiotemporal relations into its inner-working mechanisms. 

Spatiotemporal graph-based models can be divided into {hybrid} and {fused}. 
\emph{Hybrid models} combine learning algorithms developed separately for the graph domain and the temporal domain. They use graph neural networks to extract higher-level spatial features and process the latter with a temporal RNN, CNN, or variants of them. The works in \cite{chai2018bike,manessi2020dynamic,sun2020constructing} use a graph convolutional neural network (GCNN) per timestamp followed by an LSTM. Instead, authors in \cite{khodayar2018spatio} first use a temporal RNN and then a GCNN. The works in \cite{yu2017spatio,guo2019attention,wu2019graph} prefer temporal CNNs since convolutions are easier to train and have fewer parameters.
\emph{Fused models} force the graph structure into conventional spatiotemporal solutions and provide a single strategy to jointly capture the spatiotemporal relationships. They substitute the parameter matrices in these models with graph convolutional filters, which are at the core of GCNNs \cite{gama2020graphs}. The work in \cite{isufi2019forecasting} proposes a graph-based VARMA model to learn spatiotemporal representations. The works \cite{seo2018structured,si2019attention} consider the RNN family, whereas \cite{ruiz2020gated} discusses also graph-based gating \cite{Isufi20-EdgeNets}. The work in \cite{yan2018spatial} builds replicas of the spatial graph, connects nodes at time $t$ with their replicas at time $t-1$, and learns over this larger graph.

Hybrid models have the advantage that their spatial and temporal blocks are modular and can be implemented efficiently. But it remains unclear how to best interleave these blocks for learning from spatiotemporal relationships. Instead, fused models capture naturally these relationships as they have graph-time dependent inner-working mechanisms. One effective way to represent spatiotemporal relationships is to model the time as a graph (e.g., directed line); the evolution as a \emph{time-varying} signal over this graph \cite{sandryhaila13-dspg}; and the overall data as a \emph{time-invariant} signal over the product graph between the spatial and the temporal graph \cite{sandryhaila2014big}. This solution has resulted useful to develop a graph-time Fourier representation \cite{grassi2017time}, autoregressive models \cite{isufi2016autoregressive,isufi2019forecasting}, and signal interpolation algorithms \cite{romero2017kernel}. However, despite the success of product graphs to capture spatiotemporal relations, learning solutions over product graphs remain little explored.

In this paper, we develop a graph-time convolutional neural network (GTCNN) that implements a compositional learning model, where each layer performs convolutions over the product graph. Our contribution is threefold:
\begin{enumerate}[label=\textbf{C.\arabic*.}]
\item We develop a graph-time convolutional module build from the first principles of the convolution operator \cite{lecun1995convolutional,sandryhaila13-dspg} over a parametric product graph \cite{natali2020forecasting}. Working with first principles and parametric product graphs allows learning the spatiotemporal coupling from data and generalizes  \cite{yan2018spatial}, which can be seen as an order one convolutional filter over the Cartesian product.
\item We propose a recursive implementation of the graph-time convolutional module, which avoids working with large product graphs. This recursive implementation has a linear cost in the product graph dimensions and a constant number of parameters.
\item We develop a graph-time pooling module based on a zero-padding strategy \cite{gama18-gnnarchit} to reduce the number of active nodes; hence parameters. The advantage of zero-pad pooling is that it preserves the original spatial graph structure in the deeper layers and does not resort to coarsening or clustering techniques.
\end{enumerate}
%
%
%
Numerical results with synthetic and three real datasets corroborate the effectiveness of the proposed approach. \vskip-.3cm

\section{Signals over Product Graphs}
\label{sec:backg}
 
Consider an $N \times 1$ multivariate signal $\bbx_t$ collected over $T$ time instances in matrix $\bbX = [\bbx_1, \ldots, \bbx_T]$, such as sensor recordings in a sensor network. Signals in $\bbX$ have \emph{spatiotemporal} relations, which if fully-exploited serve as a powerful inductive bias to learn representations \cite{battaglia2018relational}.
When signal $\bbx_t$ has an (hidden) underlying structure, we can represent its spatial relations through a \emph{spatial} graph $\ccalG = (\ccalV,\ccalE)$ of $N$ nodes in set $\ccalV = \{1, \ldots, N\}$ and $|\ccalE|$ edges in set $\ccalE \subseteq \ccalV\times\ccalV$. Signal $\bbx_t  = [x_{t1}, \ldots, x_{tN}]^\top$ are a collection of values $x_{ti}$ residing on node $i$ at time $t$. Likewise, we can capture the temporal relations by viewing each row  $\bbx^i = [x_{1i}, \ldots, x_{Ti}]^\top$ of $\bbX$ as a graph signal over the nodes of a \emph{temporal} graph $\ccalG_T = (\ccalV_T, \ccalE_T)$ of $T$ nodes $\ccalV_T = \{1, \ldots, T\}$ and $|\ccalE_T|$ edges $\ccalE_T = (t, t^\prime)$. Set $\ccalE_T$ contains an edge if signals at time instances $t$ and $t^\prime$ are related. Examples for $\ccalG_T$ are the directed line graph that assumes signal $\bbx_t$ depends only on the former instance $\bbx_{t-1}$, the cyclic graph that accounts for periodicity, or any other graph that encodes the temporal dependencies in \bbX \cite{ortega2018graph}. We will represent graphs $\ccalG$ and $\ccalG_T$ through their respective graph shift operator matrices $\bbS \in \reals^{N \times N}$ and $\bbS_T \in \reals^{T\times T}$; e.g., adjacency, Laplacians \cite{sandryhaila13-dspg,ortega2018graph}.
 
\begin{figure}[t]
       \includegraphics[width=\columnwidth]{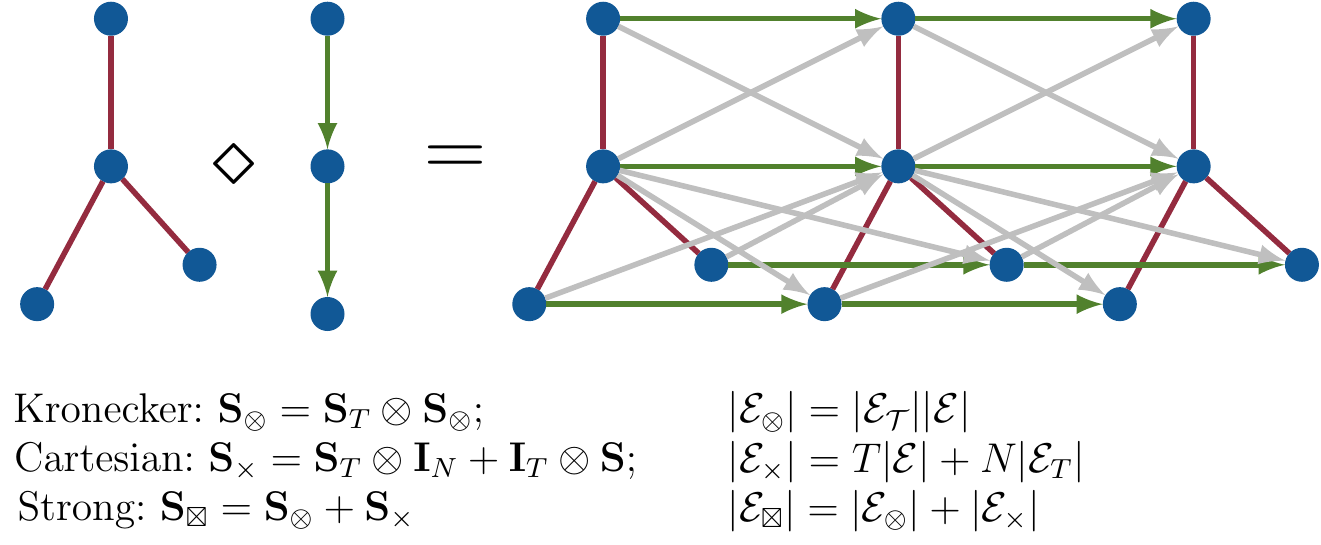}
\caption{Product Graphs. Kronecker product: $\ccalG_\otimes = \ccalG_T\otimes\ccalG$ has the grey edges. Cartesian product: $\ccalG_\times = \ccalG_T\times\ccalG$ has only the red and green edges. Strong product: $\ccalG_\boxtimes = \ccalG_T\boxtimes\ccalG$ has all edges. Parametric product: $\ccalG_\diamond = \ccalG_T \diamond \ccalG$ has all edges of the strong and self loops if all $s_{ij} \neq 0$ in \eqref{eq.paramGSO}.}\vskip-.5cm
\label{fig:Pgraphs}
\end{figure}

 Given graphs $\ccalG$ and $\ccalG_T$, we can capture the spatiotemporal relations in $\bbX$ through the product graph $\ccalG_\diamond = \ccalG_T \diamond \ccalG = (\ccalV_\diamond, \ccalE_\diamond)$, where the vertex set $\ccalV_\diamond = \ccalV_T\times\ccalV$ is the Kronecker product between $\ccalV_T$ and $\ccalV$ while the edge set depends on the product \cite{sandryhaila2014big,isufi2016autoregressive,grassi2017time,romero2017kernel, ortiz2018sampling}. Typical product graphs include the Kronecker, Cartesian, and strong product, which are particular cases of a parametric product graph with the shift operator

\begin{equation}\label{eq.paramGSO}
\bbS_\diamond = \sum_{i = 0}^1\sum_{j = 0}^1 s_{ij}\big(\bbS_T^i \otimes \bbS^j	\big)
\end{equation}
where $\{s_{ij}\}$ are scalars and $``\otimes"$ the Kronecker product \cite{natali2020forecasting}; see Fig.~\ref{fig:Pgraphs}. The parametric product graph captures the spatiotemporal coupling with the four scalars $s_{ij}$. If all $s_{ij}$s are non-zero, the parametric product graph has $|\ccalE_\diamond| = |\ccalE_{\boxtimes}| + NT$ edges, which are $NT$ mode edges than the strong product because of self-loops ($s_{00}\neq 0$).
 
 Column-vectorizing $\bbX$ yields a product graph signal $\bbx_\diamond = \text{vec}(\bbX)\in \reals^{NT}$ in which node $i_\diamond \in \ccalV_\diamond$ represents the space-time location $(i,t)$ with value $x_{ti}$, i.e., the $i_\diamond$th entry of $\bbx_\diamond$. Our goal is to exploit the coupling \emph{product graph signal}--\emph{product graph} to learn spatiotemporal representations in a form akin to temporal or graph convolutional neural networks \cite{lecun1995convolutional,gama2020graphs}.

\section{Graph-Time Convolutional Neural Networks}
\label{sec:gtcnn}

A graph-time neural network is a compositional architecture of $L$ layers each having a graph-time convolutional module, a graph-time pooling module, and a nonlinearity. 
At layer $\ell$, we have as input a collection of $F_{\ell-1}$ graph signal features $\bbx_{\diamond, \ell-1}^g $ for $g = 1, \ldots, F_{\ell-1}$. These features are the output of the previous layer and can be seen as a collection of signals over the vertices of a product graph $\ccalG_{\diamond,\ell-1}$. Each input feature $\bbx_{\diamond, \ell-1}^g$ is processed in parallel by a bank of $F_\ell$ graph-time filters $\bbH_{\ell}^{fg}(\bbS_{\diamond,\ell-1})$ to yield the aggregated (filtered) features
\begin{equation}\label{eq.filtMod}
\bbu_{\diamond,\ell}^{fg} = \bbH_{\ell}^{fg}(\bbS_{\diamond,\ell-1})\bbx_{\diamond,\ell-1}^g\quad\for\quad (f;g) = 1, \ldots, (F_{\ell}; F_{\ell-1}).
\end{equation}
Aggregated features $\bbu_{\diamond,\ell}^{fg} $ obtained from a common input $\bbx_{\diamond,\ell-1}^g$ are summed to form the higher-level linear features of layer $\ell$
\begin{equation}\label{eq.gtconvL}
\bbu_{\diamond,\ell}^f = \sum_{g = 1}^{F_{\ell-1}}\bbu_{\diamond,\ell}^{fg} = \sum_{g = 1}^{F_{\ell-1}}\bbH_\ell^{fg}({\bbS_{\diamond,\ell-1}})\bbx_{\diamond,\ell-1}^g\quad\for\quad f = 1, \ldots, F_\ell
\end{equation}
which are again a collection of $F_\ell$ product graph signals. The linear features $\bbu_{\diamond,\ell}^f $ are input in parallel to the graph-time pooling module $\rho_\ell(\cdot; \ccalG_{\diamond,\ell-1})$ to obtain the pooled features
\begin{equation}\label{eq.poolL}
\bbz_{\diamond,\ell}^f = \rho_\ell\big(\bbu_{\diamond,\ell}^f; \ccalG_{\diamond,\ell-1}\big)\quad\for\quad f = 1, \ldots, F_\ell
\end{equation}
which are vectors of dimensions $N_{\ell}T_{\ell}$ with $N_\ell \!<\! N_{\ell\!-\!1}$ and $T_{\ell} \!<\! T_{\ell\!-\!1}$ being the number of spatial and temporal nodes at layer $\ell$, respectively. Function $\rho_\ell(\cdot; \ccalG_{\diamond,\ell-1})$ signifies pooling is performed over the product graph $\ccalG_{\diamond,\ell-1}$. The pooled features are passed to a nonlinearity $\sigma(\cdot; \ccalG_{\diamond,\ell})$ to produce a collection of $F_{\ell}$ higher-level nonlinear features $\bbx_{{\diamond,\ell}}^f$ which constitute the output of layer $\ell$,
\begin{equation}\label{eq.nonlin}
\bbx_{\diamond,\ell}^f = \sigma_\ell\big[\bbz_\ell^f; \ccalG_{\diamond,\ell} 	\big]\quad\for\quad f = 1, \ldots, F_\ell
\end{equation}
where $\sigma[\cdot; \ccalG_{\diamond,\ell}]$ signifies the nonlinear function is performed over graph $\ccalG_{\diamond,\ell}$ obtained from pooling.

In the last layer $\ell = L$, we assume there is a single feature signal, which we consider the output of the graph-time neural network. We write this output compactly as
\begin{equation}\label{eq.GTCNNout}
\bbPhi\big(\bbx_\diamond; \bbS_\diamond; \ccalH	\big) = \sigma_L\bigg[\rho_L\bigg(\sum_{g = 1}^{F_{L-1}} \bbH_{L}^{g}(\bbS_{\diamond,L-1})\bbx_{\diamond,L-1}^g; \ccalG_{\diamond,L-1}	\bigg)	\bigg]
\end{equation}
to specify the dependence from the starting product graph $ \bbS_\diamond$, signal $\bbx_\diamond$, and parameters set $\ccalH$ defining all graph-time filters in \eqref{eq.gtconvL}. 

The above steps indicate that building a graph-time neural network reduces to specifying the linear filtering module $\bbH({\bbS_{\diamond,\ell}})$ [cf. \eqref{eq.filtMod}-\eqref{eq.gtconvL}], the pooling module $\rho(\cdot,\ccalG_{{\diamond,\ell}})$ [cf. \eqref{eq.poolL}], and the nonlinearity module $\sigma[\cdot;\ccalG_{\diamond,\ell}]$ [cf. \eqref{eq.nonlin}]. Each of these modules can be generalized from the corresponding ones developed for graph neural networks \cite{wu2020comprehensive}. The filtering module can be convolutional \cite{gama2020graphs}; message passing \cite{gilmer2017neural}, attention \cite{Velickovic18-GraphAttentionNetworks}, or even an edgenet-based \cite{Isufi20-EdgeNets}. Likewise, the pooling module can be zero-pad \cite{gama18-gnnarchit}, self-attention \cite{lee2019self}, or hierarchical (e.g., Kron reduction) \cite{wu2020comprehensive}. The nonlinearity module can either be pointwise (e.g., ReLU) or graph-adaptive \cite{iancu2020graph}. We will develop on a graph-time convolutional module with zero-pad pooling and pointwise ReLU nonlinearities. Our rationale is that convolutions allow for effective parameter sharing, inductive learning, and efficient implementation, while zero-pad pooling and pointwise nonlinearities make the architecture independent from graph-reduction techniques or other modules.

\subsection{Graph-Time Convolutional Filtering}

Following the shift-and-sum principle of the convolutional operator \cite{lecun1995convolutional,sandryhaila13-dspg}, we define the output of a graph convolutional filter or order $K$ over the parametric product graph $\bbS_\diamond$ as
 \begin{equation}\label{eq.grConvFilt}
 \bbu_\diamond = \sum_{k = 0}^K h_k \bbS_\diamond^k\bbx_\diamond = \sum_{k = 0}^K h_k \bigg(\sum_{i = 0}^1\sum_{j = 0}^1s_{ij} (\bbS_T^i \otimes \bbS^j)		\bigg)^k\bbx_\diamond
 \end{equation}
 where $h_0, \ldots, h_K$ are the filter parameters. Expression  \eqref{eq.grConvFilt} shifts-and-sums signal $\bbx_\diamond$ via the shift operator $\bbS_\diamond$ over the parametric product graph $\ccalG_\diamond$ to obtain the output $\bbu_\diamond$. Since $\ccalG_\diamond$ captures graph-time locations, the shifts are now performed over the spatial graph $\ccalG$ and the temporal graph $\ccalG_T$ justifying the qualifier graph-time convolution for operation \eqref{eq.grConvFilt}.\footnote{We can see the convolutional nature of \eqref{eq.grConvFilt} by particularizing the parametric product graph to the Cartesian product. Then, setting $N = 1$ node for the spatial graph with a self-loop and the temporal graph to the directed line, expression \eqref{eq.grConvFilt} implements the temporal convolution. Setting $T = 1$ node for the temporal graph with a self-loop, expression \eqref{eq.grConvFilt} reduces to a graph convolution.} Defining the graph-time filtering matrix $\bbH(\bbS_\diamond) =  \sum_{k = 0}^K h_k \bbS_\diamond^k$ allows writing \eqref{eq.grConvFilt} as  $\bbu_\diamond = \bbH(\bbS_\diamond)\bbx_\diamond$ [cf. \eqref{eq.filtMod}]. 
 
 Contrasting \eqref{eq.grConvFilt} with the graph convolutional filter \cite{gama2020graphs}, we can see that the graph-time convolutional filter aggregates at the space-time location $(i,t)$ information from space-time neighbors that are up to $K$ hops away over the product graph $\ccalG_\diamond$. This information is obtained from the shifts $\bbS_\diamond\bbx_\diamond, \ldots, \bbS_\diamond^K\bbx_\diamond$. The space-time location $(i,t)$ in $\ccalG_\diamond$ receives in this way information from other space-time locations $(j,\tau)$ that are up to $K$ hops away in $\ccalG_\diamond$ for $j \in \ccalV$ and $\tau \in [T]$. In other words, node $i$ at time $t$ receives \emph{present} signal information $\{x_{jt}\}$ from its spatial $K-$hop neighbors, and \emph{past} information $\tau < t$ from itself $\{x_{i\tau}\}$ and its spatial neighbors $\{x_{j\tau}\}$ that can be reached through a path of length $K$ in $\ccalG_\diamond$. Thus, the filter order $K$ controls the spatiotemporal locality of the graph-time convolutional filter \eqref{eq.grConvFilt}.


\smallskip
 \noindent\textbf{Computation \& recursive implementation.} We now discuss the recursive implementation of the proposed GTCNN to provide insights on its computational complexity and scalability. While working with the product graphs, we can exploit the sparsities in $\bbS_T$ and $\bbS$ to reduce the computational cost for the output \eqref{eq.grConvFilt}. If parameters $\{s_{ij}\}$ are fixed (i.e., the product graph), we can work directly with $\bbS_\diamond$, which has a sparsity of order $|\ccalE_\diamond| \!=\! NT \!+\! N|\ccalE_T| \!+\! T|\ccalE| \!+\! |\ccalE_T||\ccalE|$. Computing output $\bbu_\diamond$ requires computing the shifts $\bbx_\diamond^{(k)} = \bbS_\diamond^k\bbx_\diamond$. For this, we can use the well-know recursive implementation of shifting signals over a graph \cite{isufi2016autoregressive, gama18-gnnarchit,iancu2020graph} and write $\bbx^{(k)}_\diamond \!=\! \bbS_\diamond^k\bbx_\diamond = \bbS_\diamond\bbx_\diamond^{(k-1)}$ with $\bbx^{(0)}_\diamond = \bbx_\diamond$; hence, we can obtain the output $\bbu_\diamond$ with the linear cost $\ccalO(K|\ccalE_\diamond|)$.
 

If parameters $\{s_{ij}\}$ are to be learned (i.e., the product graph), computing $\bbS_\diamond$ beforehand or using \eqref{eq.grConvFilt} can be unaffordable in large-scale settings because of the powers of $\bbS_\diamond$ (cubic cost in $NT$). To allow scalability, we first expand all  polynomials of order $k$ and rearrange the terms to write \eqref{eq.grConvFilt} as
 \begin{equation}\label{eq.grConvFilt1}
 \bbu_\diamond = \sum_{k = 0}^K h_k \bbS_\diamond^k\bbx_\diamond = \sum_{k = 0}^{\overline{K}}\sum_{l = 0}^{\widetilde{K}}h_{kl}\big(\bbS_T^l \otimes \bbS^k	\big)\bbx_\diamond
  \end{equation}
 for some orders $\overline{K}$ and $\widetilde{K}$ and parameters $\{h_{kl}\}$. To compute output $\bbu_\diamond$, we need to compute all terms of the form $\bbx_\diamond^{(kl)} = (\bbS_T^l \otimes \bbS^k)\bbx_\diamond$. Exploiting the Kronecker product property $(\bbA\otimes\bbB)(\bbC\otimes\bbD) = \bbA\bbC \otimes \bbB\bbD$, we can write the latter as
 \begin{equation}
 \bbx_\diamond^{(kl)} = (\bbS_T\otimes\bbI_N)(\bbI_T \otimes \bbS) (\bbS_T^{l-1} \otimes \bbS^{k-1})\bbx_\diamond.
 \end{equation}
 Thus, we can compute $ \bbx_\diamond^{(kl)} $ again recursively as
 \begin{align}\label{eq.recKron}
 \begin{split}
  \bbx_\diamond^{(kl)} &= (\bbS_T\otimes\bbI_N)(\bbI_T \otimes \bbS) \bbx_\diamond^{(k-1,l-1)}\\
  &= (\bbS_T\otimes\bbI_N) \bbx_\diamond^{(k,l-1)}
  \end{split}
 \end{align}
with initialization $\bbx_\diamond^{(00)} = \bbx_\diamond$. Recursion \eqref{eq.recKron} implies we can compute $\bbx_\diamond^{(kl)}$ from $\bbx_\diamond^{(k-1,l-1)}$ with a cost of order $\ccalO(T|\ccalE| + N|\ccalE_T|)$ and since we need to perform the latter for all $k \in [\overline{K}]$ and $l \in [\widetilde{K}]$, we have a computational cost of order $\ccalO(\overline{K}T|\ccalE| + \widetilde{K}N|\ccalE_T|)$, which is linear in the product graph dimensions.

Note also that form \eqref{eq.grConvFilt1} improves our control on the spatiotemporal locality through orders $\overline{K}$ and $\widetilde{K}$. A larger $\overline{K}$ implies more reach over the spatial graph (i.e., $\bbS^{\overline{K}}$), while a larger $\widetilde{K}$ implies more reach over the temporal graph  (i.e., $\bbS_T^{\widetilde{K}}$). Both orders are design choices.

\subsection{Graph-Time Pooling}\label{sec.gtpool}

Building upon \cite{gama18-gnnarchit}, we propose a zero-pad graph-time pooling module to reduce the dimensionality of the graph-time features without resorting to any coarsening approach. This simple, yet non-trivial, generalization needs to account now for the spatiotemporal peculiarities induced by the product graph. The pooling approach has three steps: $i)$ summarization; $ii)$ slicing; $iii)$ downsampling.

\smallskip
 \noindent\textbf{Summarization} changes the signal value of a node with a summary (e.g., max, mean) of the values in the local neighborhood. Given the $N_{\ell-1}T_{\ell-1}$ convolutional features $\bbu_{\diamond,\ell-1}$, graph $\bbS_{\diamond,\ell-1}$, and defined the reachability integer $\alpha_\ell$, we denote the summarized features as
 \begin{equation}\label{eq.summariz}
 \bbv_{\diamond,\ell} = \Gamma(\bbu_{\diamond,\ell-1}; \alpha_\ell; \bbS_{\diamond,\ell-1})
 \end{equation}
 which signifies the $i$th entry $[\bbv_{\diamond,\ell}]_i$ is the summarization of signal values $[\bbu_{\diamond,\ell-1}]_j$ from nodes $j$ that are up to $\alpha_\ell$ hops away, i.e., $\{ j: [\bbS_{\diamond,\ell-1}^k]_{ij} \neq 0~\text{for some}~k \le \alpha_\ell\}$. Function $\Gamma(\cdot)$ can be, for instance, max$(\cdot)$ or mean$(\cdot)$. Since the local neighborhood of a node in $\ccalG_{\diamond,\ell-1}$ includes also spatial nodes from different time instances, the features in $ \bbv_{\diamond,\ell} $ are summarized over both the graph and the temporal domain. The $N_{\ell-1}T_{\ell-1}$ summarized features in $ \bbv_{\diamond,\ell} $ can now be seen as another product graph signal over graph $\ccalG_{\diamond,\ell-1}$.  
 
 \begin{remark} Summarization is an implicit low-pass operation and the type of product graph has an impact on its severity. If the product is parametric or strong, summarization is performed over larger spatiotemporal neighborhoods due to the inter-connections in different time instances [cf. Fig.~\ref{fig:Pgraphs}; grey edges]. This wide summarization leads to a stronger low pass and reduces the signal variability. We limit the local spatiotemporal neighborhood by using the Cartesian product [cf. Fig.~\ref{fig:Pgraphs}; green edges] in the summarization step. \hfill$\blacksquare$
 %
 %
  \end{remark}
  
  \smallskip
 \noindent\textbf{Slicing} reduces the dimensionality across the temporal dimension. Given the $N_{\ell-1}T_{\ell-1}$ summarized features $\bbv_{\diamond,\ell}$, we de-vectorize it into the union of $T_{\ell-1}$ spatial graph signals
\begin{equation*}
\big[\bbv_{\ell,1},\ldots,\bbv_{\ell,T_{\ell-1}}	\big] =  \text{vec}^{-1}\big({\bbv_{\diamond,\ell}}\big)
\end{equation*}
where each $\bbv_{\ell,t}$ is a spatial graph signal of dimension $N_{\ell-1}$. Denoting the slicing ratio at layer $\ell$ by $R_\ell$, we keep from $\big[\bbv_{\ell,1},\ldots,\bbv_{\ell,T_{\ell-1}}	\big]$ one column (or slice) every $R_\ell$, resulting in the $T_\ell = \lceil T_{\ell-1}/R_\ell\rceil$ output slices $\{\bbw_{\ell,\tau}\}$. These output slices are vectorized back into the product graph signal $\bbw_{\diamond,\ell} = \text{vec}\big(\big[\bbw_{\ell,1},\ldots,\bbw_{\ell,T_{\ell}}	\big]\big)$. Denoting the slicing operation at layer $\ell$ as $\Delta_\ell(\cdot): \reals^{N_{\ell-1}T_{\ell-1}} \to \reals^{N_{\ell}T_\ell}$, we can write the sliced features as
\begin{equation}\label{eq.slicing}
\bbw_{\diamond,\ell} = \Delta_\ell\big(\bbv_{\diamond,\ell} ;R_\ell	\big).
\end{equation}
I.e., slicing a product graph signal $\bbv_{\diamond,\ell}$ of dimensions $N_{\ell-1}T_{\ell-1}$ yields another product graph signal of dimensions $N_{\ell-1}T_{\ell}$, in which only the temporal dimension is reduced. The product graph over which the sliced signal $\bbw_{\diamond,\ell}$ resides can be built using the same rule we built the initial product graph $\ccalG_{\diamond}$ but with a smaller temporal graph of $T_{\ell} \le T_{\ell-1}$ nodes (e.g., a directed line containing fewer nodes).


\smallskip
\noindent\textbf{Downsampling} reduces the number of active nodes across the spatial dimension from $N_{\ell-1}$ to $N_{\ell}$ without modifying the underlying spatial graph. This is done via zero-padding, i.e., we set to zero the value of inactive nodes while preserving the value on the active ones. Given a binary sampling matrix $\bbC_\ell$ from the combinatorial set
\begin{equation*}
\ccalC_\ell = \{\bbC_\ell \in \{0,1\}^{N_\ell T_\ell \times N_{\ell-1}T_{\ell-1}}	: \bbC_\ell\bbone = \bbone, \bbC_\ell^\top\bbone \preceq \bbone\}
\end{equation*}
we compute the downsampled features as
\begin{equation}\label{eq.donwsampl}
\bbz_{\diamond,\ell} = \bbC_\ell \bbw_{\diamond,\ell}.
\end{equation}
When $[\bbC_\ell]_{ij} = 1$, it means the $j$th component of $\bbw_{\diamond,\ell}$ is selected and stored in the $i$th entry of $\bbz_{\diamond,\ell}$. Therefore, vector $\bbz_{\diamond,\ell}$ is a product graph signal of dimensions $N_\ell T_\ell$ residing over the active nodes $N_{\ell-1}T_{\ell}$ of the product graph $\ccalG_{\diamond,\ell}$ obtained from the slicing step. The remaining nodes in $\ccalG_{\diamond,\ell}$ have a zero value.

Downsampling through the sampling matrix $\bbC_\ell$ requires discussing two main aspects. First, designing $\bbC_\ell$ is to a large extent an art and it needs to capture the physicality of the problem but also the coupling between the higher-level features at layer $\ell$ and the spatial graph. We consider $\bbC_\ell$ to select the nodes with the highest degree as \cite{gama18-gnnarchit}. Second, the set of active nodes in deeper layers is by definition a subset of the active nodes in the earlier layers. To track the location of these active nodes we can consider the nested sampling matrix $\bbD_\ell = \bbC_\ell\ldots\bbC_1$ and use it for zero-padding. The sparsity of $\bbC_\ell$ and $\bbD_\ell$ can be in turn used to compute the graph-time convolutional filter output [cf. \eqref{eq.grConvFilt}] with a reduced shift operator like for the conventional zero-pad pooling in GNNs; refer to \cite[Sec. III-A]{gama18-gnnarchit} for the technicalities of this implementation.

In summary, a GTCNN is an architecture in which each layer is composed of a graph-time convolutional module defined by \eqref{eq.grConvFilt}-\eqref{eq.grConvFilt1}, a graph-pooling module defined by \eqref{eq.summariz}-\eqref{eq.slicing}-\eqref{eq.donwsampl}, and a nonlinearity module \eqref{eq.nonlin}. The higher-level features of layer $\ell$ are input to the successive layer as per \eqref{eq.filtMod} and propagated down the cascade until the final GTCNN output [cf. \eqref{eq.GTCNNout}] is obtained, which is the joint graph-time embedding. The embedded features are fed optionally to fully connected layers and then to a loss function to learn the parameters in $\ccalH = \{\{h_{k\ell}^{fg}\}; \{s_{ij,\ell}\}\}$ comprising the coefficients of all convolutional filters $\{h_{k\ell}^{fg}\}$, all parametric product graphs $\{s_{ij,\ell}\}$, and optionally of the fully connected layers. Remark the parameters in $\ccalH$ are independent of the graph dimensions, while the cost of the GTCNN is linear and governed by the graph-time convolutional filters.

\section{Numerical Results}
\label{sec:numRes}
This section corroborates the performance of the GTCNN to provide insights on its inner-working mechanisms and compare it with baseline and state-of-the-art alternatives. We used ADAM to train all models with decaying factors $\beta_1 = 0.9$ and $\beta_2 = 0.999$ \cite{kingma17-adam}. The temporal graph is the directed line.\footnote{Code available at \url{https://github.com/gtcnnpaper}.}

\begin{figure}[!t]
       \includegraphics[width=\columnwidth]{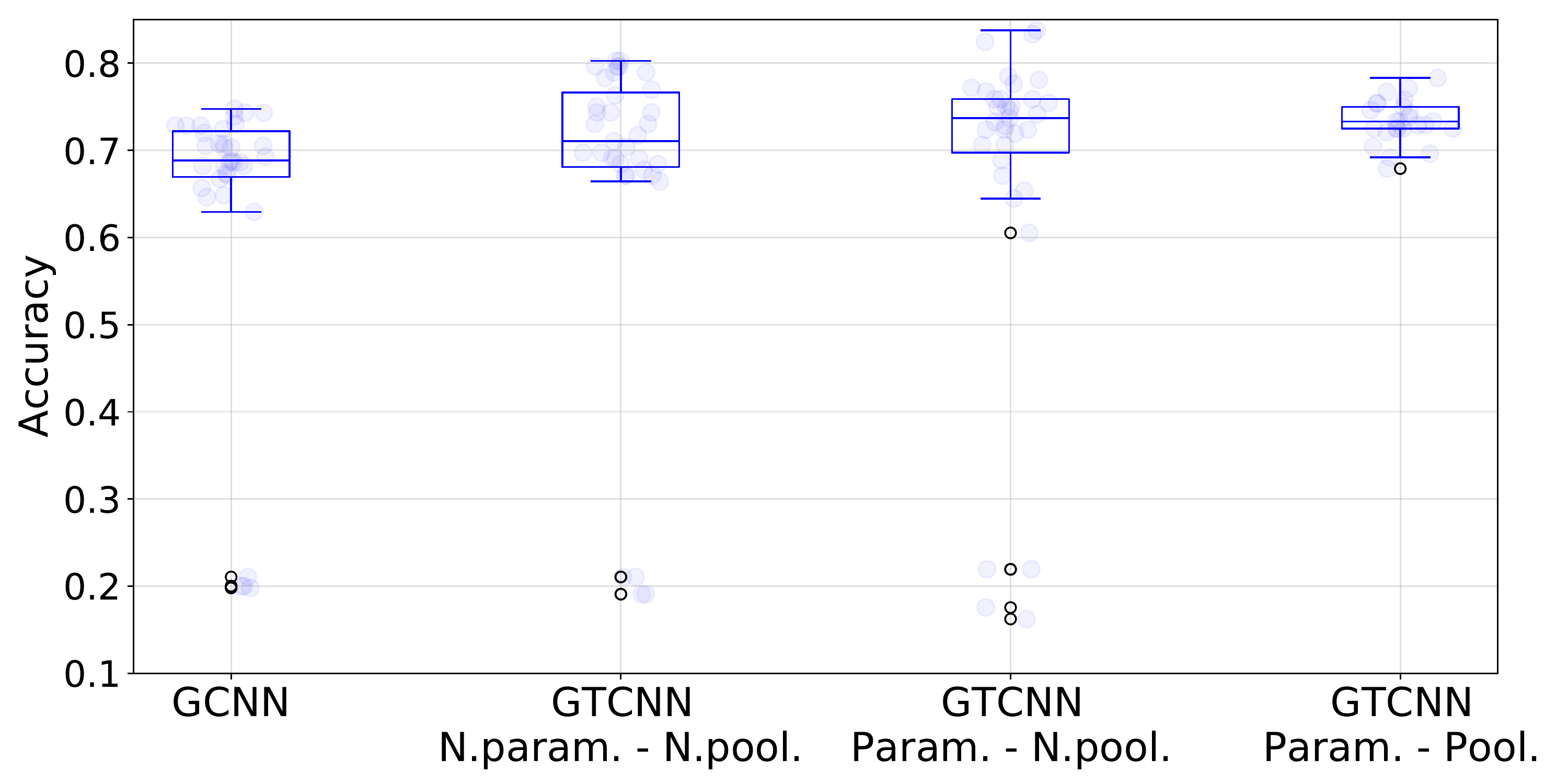}
\caption{Comparison of the GCNN baseline with the non-parametric and parametric GTCNN without pooling and with the parametric GTCNN with pooling.}\vskip-.4cm
\label{fig:GTCNNPerf}
\end{figure}


\subsection{Source Localization}\label{sec.SourcLoc}

First, we consider a controlled synthetic experiment to highlight the role of the GTCNN key components. The task consists of finding the source of a diffusion process by observing a sequence of $T$ graph signals $\bbx_t, \ldots,\bbx_{t+T}$ for a random time instance $t$. The graph is an undirected stochastic block model of $N = 100$ nodes and $C = 5$ communities. The basic experimental setup is the same as in \cite{gama18-gnnarchit} but we considered $1,200$ data points to avoid duplicates with an $80\%-10\%-10\%$ split. For a fair comparison with the graph-only GCNN baseline, we considered $T$ successive signal realizations as features in the input layer. All architectures have two layers and two filters per layer of order two and are trained over $8,000$ epoch with a batch size of $100$ samples. We evaluated features in $F_1, F_2 \in \{2, 4, 16, 32\}$, downsampling sizes $N_1 \in \{30, 75, 100\}$ and $N_2 \in \{10, 30, 50\}$, and temporal windows $T \in \{1, 2, 3\}$; see [Supplement;~Sec.~I].

Fig.~\ref{fig:GTCNNPerf} compares the GTCNNs with parametric and non-parametric product graphs [cf. Fig.~\ref{fig:Pgraphs}], with and without pooling, and with the baseline GCNN. We can see that accounting for the temporal domain via the sparse connectivity of the product graph improves upon GCNN solutions. Better results are achieved via the parametric product graph and by the use of pooling as evidenced by the larger median value and the smaller deviation of the right-most boxplot. Differently from the others, the latter architecture has also no negative outliers, which indicates it learned in all graphs and data splits. We attribute the latter to the fact that the spatiotemporal coupling is learned in a sparse way and to the zero-pad pooling that preserves the original spatial graph.

\subsection{Forecasting}\label{sec.tempForec}

We now consider the task of forecasting future values of a multivariate time-varying signal given a sequence of $T$ past realizations. We used the setting in \cite{isufi2019forecasting} and considered the Molene dataset comprising $744$ hourly temperature measurements across $N = 32$ stations in a region of France; and the NOAA dataset comprising $8,579$ hourly temperature measurements across $109$ in the U.S.. The loss function is the MSE between the one-step ahead prediction $\hbx_{t+1}$ and the true value $\bbx_{t+1}$ regularized by the norm-one of all parametric product graph coefficients $\bbs = \text{vec}(\{s_{ij,\ell}^{fg}\})$ [cf. \eqref{eq.paramGSO}], i.e., $\ccalL = \text{MSE}(\hbx_{t+1}; \bbx_{t+1}) + \beta \|\bbs\|_1$, where $\beta \ge 0$ is a scalar. We compared the GTCNN with: $i)$ the linear models G-VARMA and GP-VAP \cite{isufi2019forecasting}; $ii)$ the gated graph-based RNN (GGRNN) \cite{ruiz2020gated}; $iii)$ the time-only LSTM. For the G-VARMA and GP-VAP we used the parameters from \cite{isufi2019forecasting}, while for the GTCNN and GGRNN we evaluated features $F \in \{2, \ldots, 20\}$, orders $K \in \{2, \ldots, 5\}$, observation windows $T \in \{3, 4, 5\}$, and norm-one sparsity weights $\beta \in [0, 0.05]$. For the LSTM we varied the number of hidden units in $\{8, 16, 32, 64\}$. We also considered different learning rates in $[5\times 10^{-4},10^{-3}]$.

\begin{figure}[t]
       \includegraphics[width=\columnwidth]{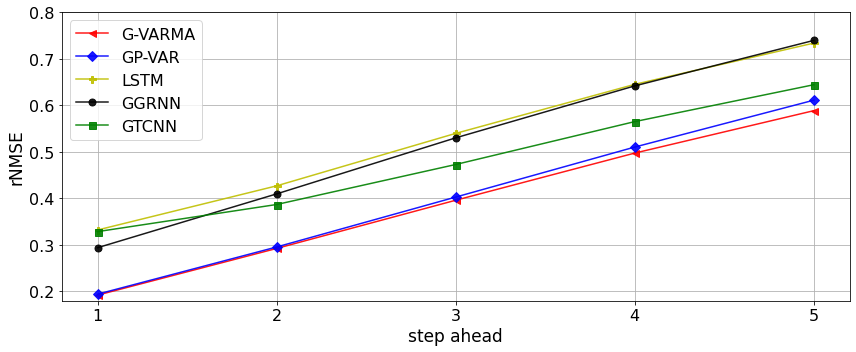}\vskip-.1cm
\caption{Root normalized MSE versus future prediction steps for the different methods in the Molene dataset.}\vskip-.3cm
\label{fig:MoleneComp}
\end{figure}

\begin{figure}[t]
       \includegraphics[width=\columnwidth]{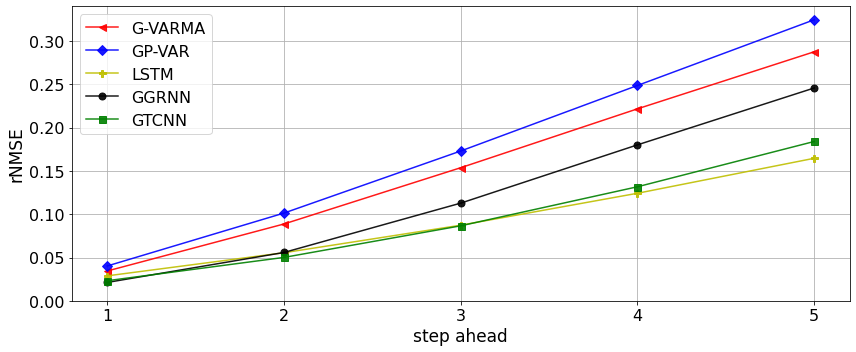}\vskip-.1cm
\caption{Root normalized MSE versus future prediction steps for the different methods in the NOAA dataset.}\vskip-.4cm
\label{fig:NoaaComp}
\end{figure}

Figs.~\ref{fig:MoleneComp} and~\ref{fig:NoaaComp} show the root normalized MSE (rNMSE) for up to five steps ahead prediction for the Molene and the NOAA dataset, respectively. For the Molene dataset, we can see that all graph-based approaches achieve a lower rNMSE than the LSTM. This is because the dataset contains fewer training samples; thus, imposing an inductive bias \cite{battaglia2018relational} through the product graph during learning is helpful. In fact, the best performance is achieved by the linear graph models, while the GTCNN performs the best among the neural network alternatives. For the NOAA dataset, instead, we see the opposite trend: the neural network solutions achieve a lower rNMSE compared with the linear graph models. Since the NOAA dataset contains more training samples it allows neural networks to learn more complicated representations. The GTCNN achieves the best performance together with the LSTM, while the GGRNN suffers when predicting more than three steps. Overall, these results put the GTCNN as a valid alternative to learn representations with inductive spatiotemporal biases when both the number of training samples is limited and large.

\subsection{Earthquake classification}
Lastly, we propose an experiment to find the epicenter of precursor-based earthquakes \cite{geller1997earthquake}. Precursor-based detection relies only on wave recordings up to $20$ seconds before the strike but not on historical trends. This is a challenging task and our main goal is to show how the GTCNN can be used to approach the latter. We built a dataset from the New Zeeland earthquake service (Supplement~Sec.~II).

\begin{figure}[t]
      \qquad \includegraphics[width=0.4\columnwidth]{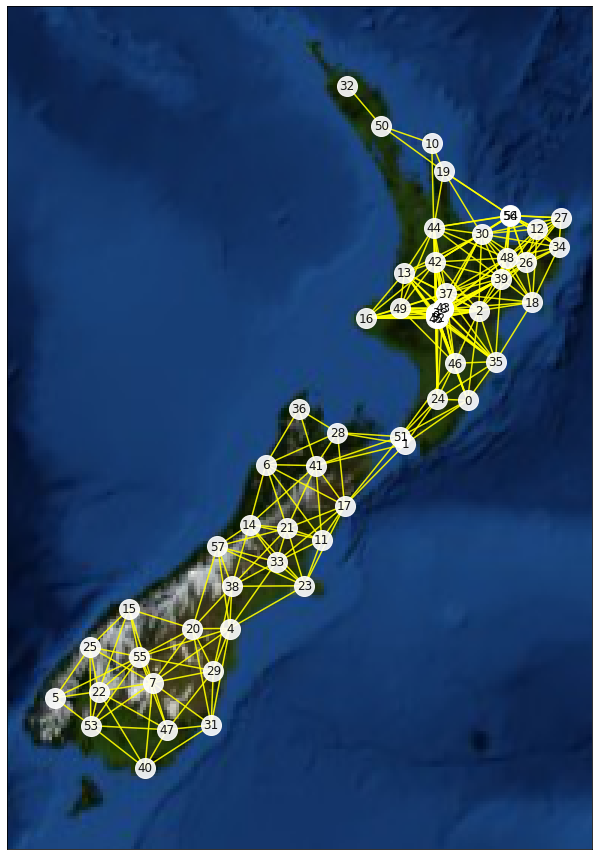}\qquad
       \includegraphics[width=0.4\columnwidth]{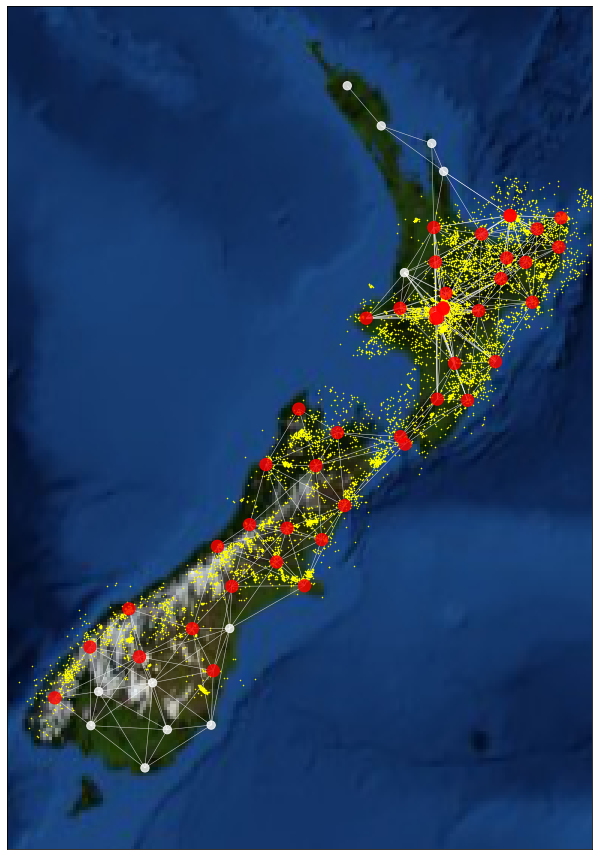}\hfill
\caption{(Left) Graph structure among the seismic stations. (Right) Yellow dots are the earthquake epicentres; red nodes are stations with an assigned label; white nodes are stations without a label.}\vskip-.3cm
\label{fig:EQ-graph}
\end{figure}

\smallskip
\noindent\textbf{Experimental setup.} We considered $4,633$ seismic wave recordings between 2016 and 2020 across $58$ stations. These recordings are of earthquakes with a magnitude between one and three and not further than $200$km from the closest station. The epicenter is assigned to one of the seismic stations and in total there are $45$ stations with assigned labels each with approximately $175$ earthquakes. The remaining $13$ stations record the waves but do not have an assigned earthquake. We built a geometric graph of $N = 58$ nodes and an edge exists if two stations are within $170,3$km; like the Molene/NOAA graphs in \cite{isufi2019forecasting}. Fig.~\ref{fig:EQ-graph} illustrates the graph and the earthquake distribution. The graph signal consists of $20$ timesteps of recording in the ten seconds before the strike over all $58$ stations. We compared again the GTCNN with the LSTM and the GGRNN. All models are trained w.r.t. the cross-entropy loss for $100$ epochs with a batch size of $128$, learning rate $10^{-3}$, and early stopping at $20$ epochs. The results are averaged over $20$ realizations. The GTCNN has three layers with grid-searched features $F_1 = 4, F_2 = 8, F_3 = 12$, filter orders $K = 2$, slicing ratios $R = 2$, and active pooling nodes $100\%$, $90\%$, and $70\%$ of the total nodes in layers one, two, and three, respectively. The LSTM has $20$ grid-searched hidden units, while the GGRNN has the same parameters as in \cite{ruiz2020gated}. We considered two experiments: \emph{one-vs-all} binary classification, which assigns the wave to a specific station or any of the other $44$ stations; \emph{all-vs-all} $45$ class classification scenario, which assigns the wave to one of the $45$ stations.

\begin{figure}[!t]
       \includegraphics[width=\columnwidth]{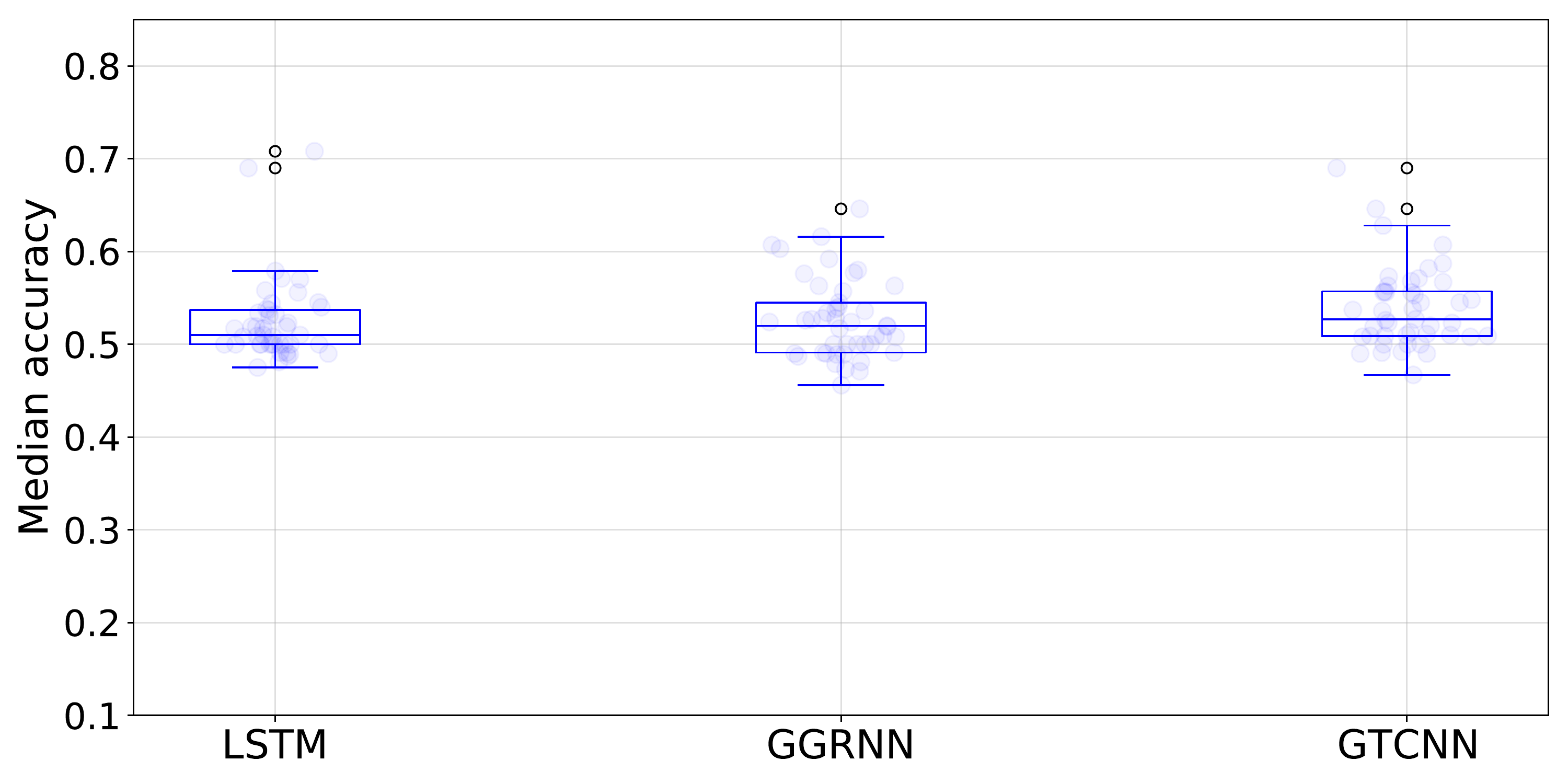}
       \caption{Median accuracy distribution of the different stations in the one-vs.-all experiment. We can see the GGRNN and the GTCNN have more stations achieving median accuracies of more than $60\%$, where the GTCNN has a few stations exceeding $65\%$ of accuracy.}
\label{fig:EQ-medianAccruaciesl}\vskip-.5cm
\end{figure}

\begin{table}[!t]
\centering
\caption{Performance of the different methods for the one-vs-all earthquake classification task. In brackets, it is shown the standard deviation of the respective metric. See supplement for each class.}
\begin{tabular}{lcccc}
    \hline\hline
    {Model}                                                     & {Accuracy} & {Precision}   & {Recall}       & {F1}                                  \\ \hline
      \rowcolor{Gray}
{LSTM}                                                     & 0.52 (0.05) &  0.53 (0.05)   &  0.53 (0.06) & 0.50 (0.06)                                       \\ 
{GGRNN \cite{ruiz2020gated}}                                                     & 0.53 (0.04) & 0.53 (0.05) &  0.54 (0.06) & 0.51 (0.05)  \\
  \rowcolor{Gray}
{GTCNN}                                                     & 0.54 (0.04) & 0.54 (0.05)   &  0.55 (0.06) & 0.53 (0.05)                                       \\ 
\hline\hline
\end{tabular}
\label{tab.EQbinary}
\end{table}

\smallskip
\noindent\textbf{One-vs-all:} In this setting, we balanced the dataset by considering half of the points from the class of interest and the other half from all remaining $44$ stations. From Fig.~\ref{fig:EQ-medianAccruaciesl} and Table~\ref{tab.EQbinary}, we can see that, while all methods have a very similar statistical performance, the GTCNN has a higher average value compared with the other alternatives.

\begin{figure}[t]
       \includegraphics[width=\columnwidth]{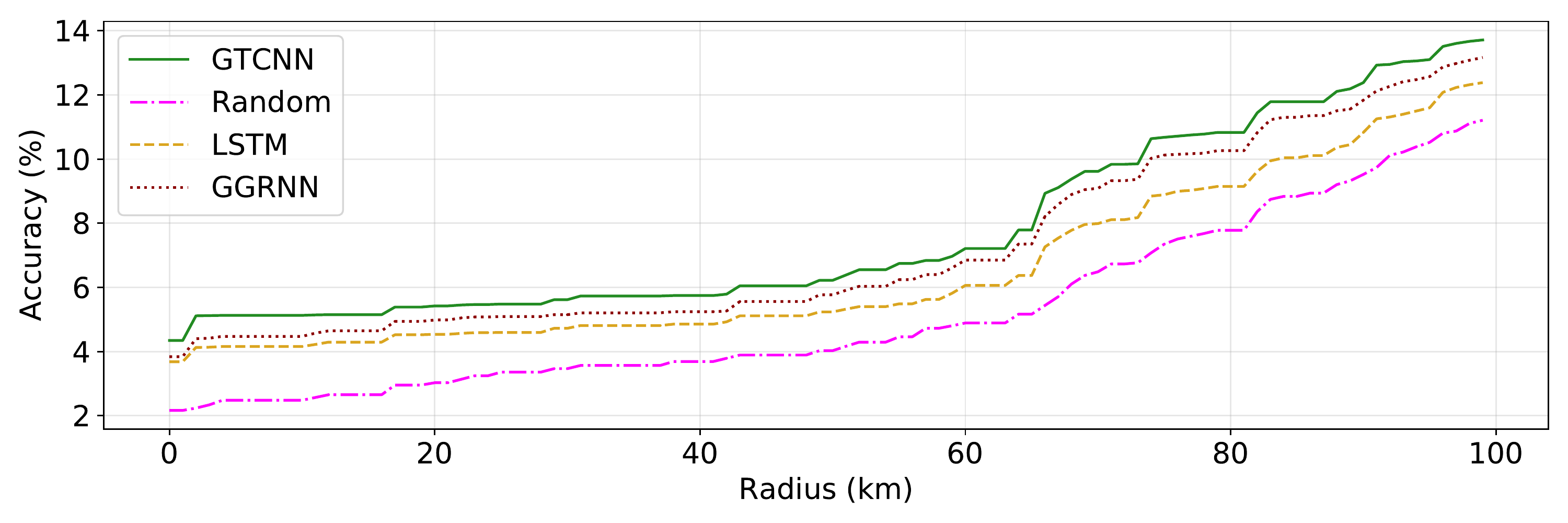}
       \caption{Radius-based accuracy results of the different models as a function of the distance from the correct station. I.e., a classification for an earthquake with label station $i$ is considered correct even if assigned to a station that is within a given radius.}\vskip-.5cm
\label{fig:EQ-all}
\end{figure}

\smallskip
\noindent\textbf{All-vs-all:} Since for the approximately $4,5$k data points a $45$ class classification problem is challenging, we measure the performance with a radius-based accuracy metric. That is, if an earthquake has as correct label station $i$, we consider a correct classification also a station within the radius. Another reason for such a choice is that several earthquakes have their epicenters far from any station (e.g., $70$km) or between two or more stations. We considered a split of $60\%-20\%-20\%$. From Fig.~\ref{fig:EQ-all}, we can see that graph-based methods perform better than the LSTM highlighting again the impact of this prior when the problem is challenging. As we increase the radius, the performance of all approaches increases with the GTCNN achieving a slightly better result. Note that even by increasing the radius just to $3$km the GTCNN shows the biggest jump, which indicates it has assigned several epicenters to stations close to the true label.

We may still correctly argue the reported performance is still far from satisfactory. Reasons for these are multiple (station distribution across the country; the match between the spatial graph and wave propagation; use of only the vertical velocity of the wave), but they, however, show promise for the GTCNN and the other graph-based solutions; and in the one-vs-all accuracy (Fig.~\ref{fig:EQ-medianAccruaciesl}) we have also seen accuracies for particular stations up to $80\%$; see Supplement~Sec. II.

\section{Conclusions}
\label{sec:conclusions}

We proposed a graph-time convolutional neural network to learn spatiotemporal dependencies with a convolutional prior over both the graph and temporal domain. The spatiotemporal data are first transformed into a static signal over a larger product graph between the spatial relationship graph and the temporal relationship graph. The product graph is parametric such that we can learn the spatiotemporal coupling directly from the data. The convolutional module follows the first principles of the convolution operator and builds the output as a shift-and-sum of the input signal over the product graph. A graph-time pooling module is proposed based on spatial zero-padding to preserve the spatial graph-prior in the deeper layers and with a temporal slicing to reduce the dimension across time. We corroborated the GTCNN on classification and regression tasks showcasing its ability to learn spatiotemporal representations. By providing a new alternative to learn from temporal data, the GTCNN opens the doors to a novel research stream including applications from different scientific disciplines. Future work will consider parallelization of the GTCNN to handle large-scale graphs and theoretical advances to shed light on the capability of the GTCNN to discriminate graphs in a spatiotemporal manner.

\newpage
\bibliographystyle{IEEEtran}
\bibliography{myIEEEabrv,bib-nonlinear}
\newpage

\subsection*{Supplementary material}\label{sec.supp}

\setcounter{figure}{0}
\setcounter{table}{0}
\setcounter{section}{0}

\noindent This document contains the supplementary material of the paper \emph{Graph-Time Convolutional Neural Networks}. Section~\ref{sec.suppSouLoc} analyzes of the different graph convolutional neural network (GTCNN) components in the source localization dataset. Section~\ref{sec.suppEQ} provides details about the earthquake labeling experiment and the dataset analysis.\vskip-.2cm
%
\section{Source localization}\label{sec.suppSouLoc}\vskip-.3cm
In this section, we analyze the impact of the type of product graph and pooling on the GTCNN.

\smallskip
\noindent\textbf{Product graph.} We start with a GTCNN of two layers of $F_1 = F_2 = 2$ and two filters per layer of orders $K = 2$. Our rationale is that this minimalistic architecture allows understanding better how much the product graph aids learning: $i)$ considering more features per layer will lead to a more complex network that may overfit; and $ii)$ the results are independent of pooling, which we analyze next.

\begin{table}[!h]
\centering
\caption{Mean accuracy (std. dev.) of the GTCNN with different product graphs compared with the baseline GCNN.}
\begin{tabular}{lccc}
    \hline\hline
    {Model}                                                     & {T = 1} & {T = 2}   & {T = 3}                                         \\ \hline
      \rowcolor{Gray}
{GCNN (baseline)}                                                     & {0.64 ($ 0.16$)} &  {0.42 ($ 0.21$)}   &  {0.44 ($ 0.28$)}                                         \\ 
{GTCNN Cartesian}                                                     & n/a &  {0.65 ($ 0.19$)}   &  {0.66 ($ 0.19$)}  \\
  \rowcolor{Gray}
{GTCNN Strong}                                                     & n/a &  {0.63 ($ 0.20$)}   &  {0.67 ($ 0.17$)}                                         \\ 
{GTCNN Parametric}                                                     & n/a &  {0.68 ($ 0.18$)}   &  {0.69 ($ 0.20$)}                                         \\ 
\hline\hline
\end{tabular}
\label{tab.prodGraphsRes}
\end{table}

We ignored the GTCNN with the Kronecker product since this architecture does not have a connected product graph. As a baseline, we considered a GCNN working with the time-varying signal of window $T$ as multiple features over the nodes. This comparison is shown in Table~\ref{tab.prodGraphsRes}. The parametric GTCNN achieves the highest mean performance but there is no significant difference with the strong product GTCNN. These two results suggest the temporal relations between neighboring nodes aid learning. Contrarily, if the product graph is not used and the baseline graph convolutional neural network (GCNN) is employed, we see the performance degrades substantially. This is because such a network fails more often to learn from particular graph realizations and data splits, especially, when $T\ge2$. We attribute the latter to the fact that this procedure is not exploiting the physicality of the problem to capture spatiotemporal relations in a sparse manner but rather treats them as a union of features. 

\begin{figure}[!t]
\centering
       \includegraphics[width=.9\columnwidth]{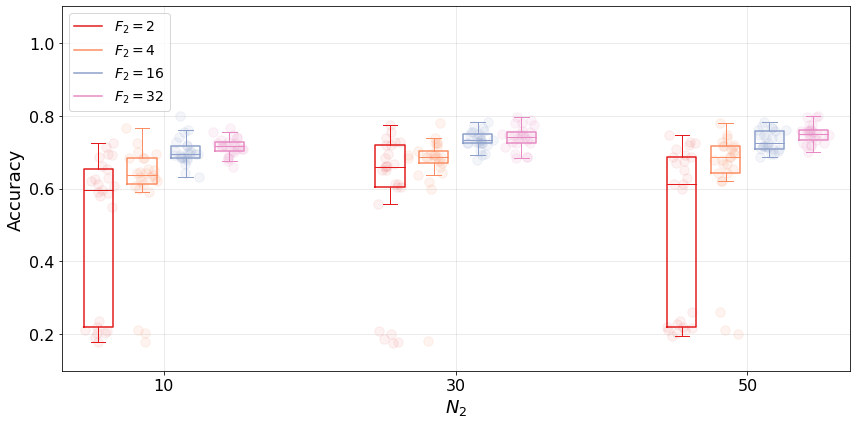}\vskip-.2cm
\caption{Accuracy versus the number of active nodes in the second layer of the GTCNN for different features. The first layer has $F_1 = 2$ features and $N_1 = 100$ active nodes. The GTCNN performs better when its expressive power (higher $F_2$) increases and requires fewer active nodes in the second layer (lower $N_2$).}\vskip-.45cm
\label{fig:GTCNNPoolL2}
\end{figure}

\smallskip
\noindent\textbf{Pooling.} We now investigate the effects of pooling in the GTCNN. We considered the two-layered parametric GTCNN with a temporal window $T = 2$. The temporal slicing ratios are $R_1 = 1$ and $R_2 = 2$, i.e., all instances are kept in the first layer and only half in the second layer. Initially, we analyze the pooling effects in the second layer for a different number of features $F_2 \in \{2, 4, 6, 16, 32\}$, and active nodes $N_2 \in \{10, 30, 50\}$. From Fig.~\ref{fig:GTCNNPoolL2}, we can see the highest performance is achieved when $F_2 \ge 16$ and $N_2 \ge 30$. This indicates that, when the GTCNN is equipped with a higher expressive power (more filters), it can allow for a more drastic pooling in the second layer without affecting the performance. Remark in the latter setting, the GTCNN has also fewer outliers (i.e., cases where it cannot learn), indicating more robustness to graph realizations and data splits.

\begin{figure}[!t]
\centering
       \includegraphics[width=.9\columnwidth]{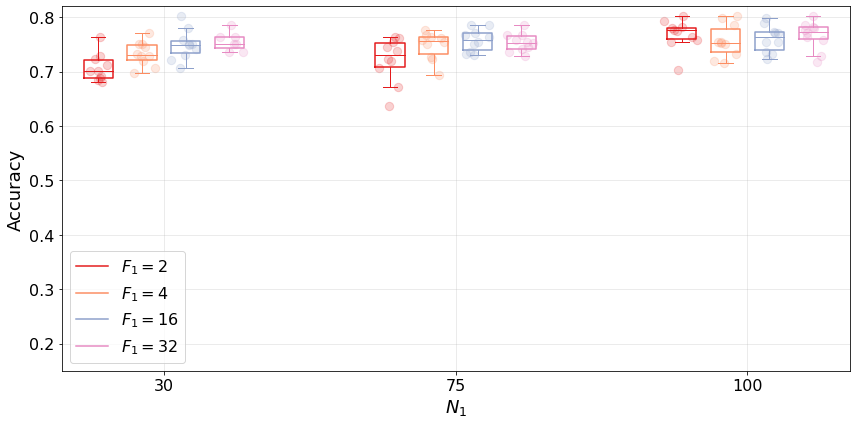}\vskip-.2cm
\caption{GTCNN performance for different downsampling nodes and features in the second layer.}\vskip-.45cm
\label{fig:GTCNNPoolL1}
\end{figure}

Next, we analyze the effects of pooling in the first layer. From the earlier results, we fix $F_2 = 16$ features and $N_2 = 30$ active nodes and test for $F_1 \in \{2, 4, 16, 32\}$ and $N_1 \in \{30, 75, 100\}$ to have $N_1 \ge N_2$. From Fig.~\ref{fig:GTCNNPoolL1}, we can see the performance degrades when more filters are considered in the first layer. This is because the network with $F_2 = 16$ filters has already sufficient discriminatory power for the dataset at hand and increasing it further leads to overfitting. From the pooling perspective, these results indicate that all nodes ($N_1 = 100$) should be kept in the first layer (higher median and lower spread). This is not entirely surprising since the first layer learns lower-level representations and exploits all data. The latter observation is particularly visible for $F_1 = 2$ when the network is less prone to overfitting.
In conclusion, these results indicate the GTCNN may require all lower level features in the input layer to learn the intermediate representation but can sacrifice a large portion of active nodes in the next layer without affecting the performance.

\section{Earthquake classification}\label{sec.suppEQ}\vskip-.2cm

\begin{figure}[t]
\centering
       \includegraphics[width=0.65\columnwidth]{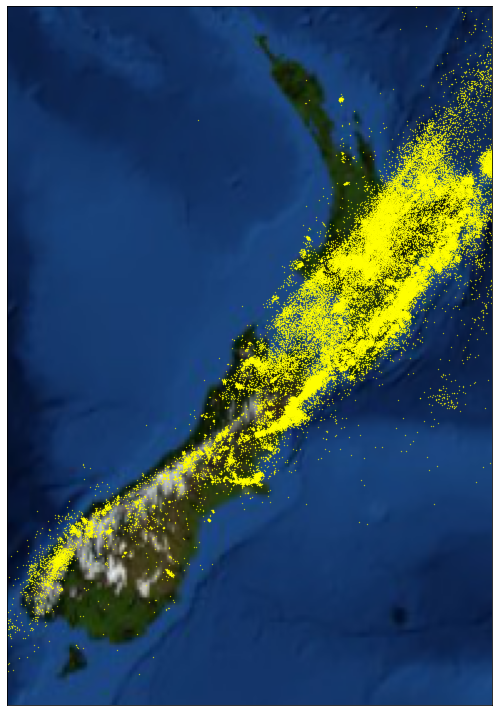}
\caption{ Earthquake Epicentre Distribution. Each yellow point represents an earthquake.}\vskip-.45cm
\label{fig:EQdistrib}
\end{figure}

\begin{figure}[t]
\centering
       \includegraphics[width=0.49\columnwidth]{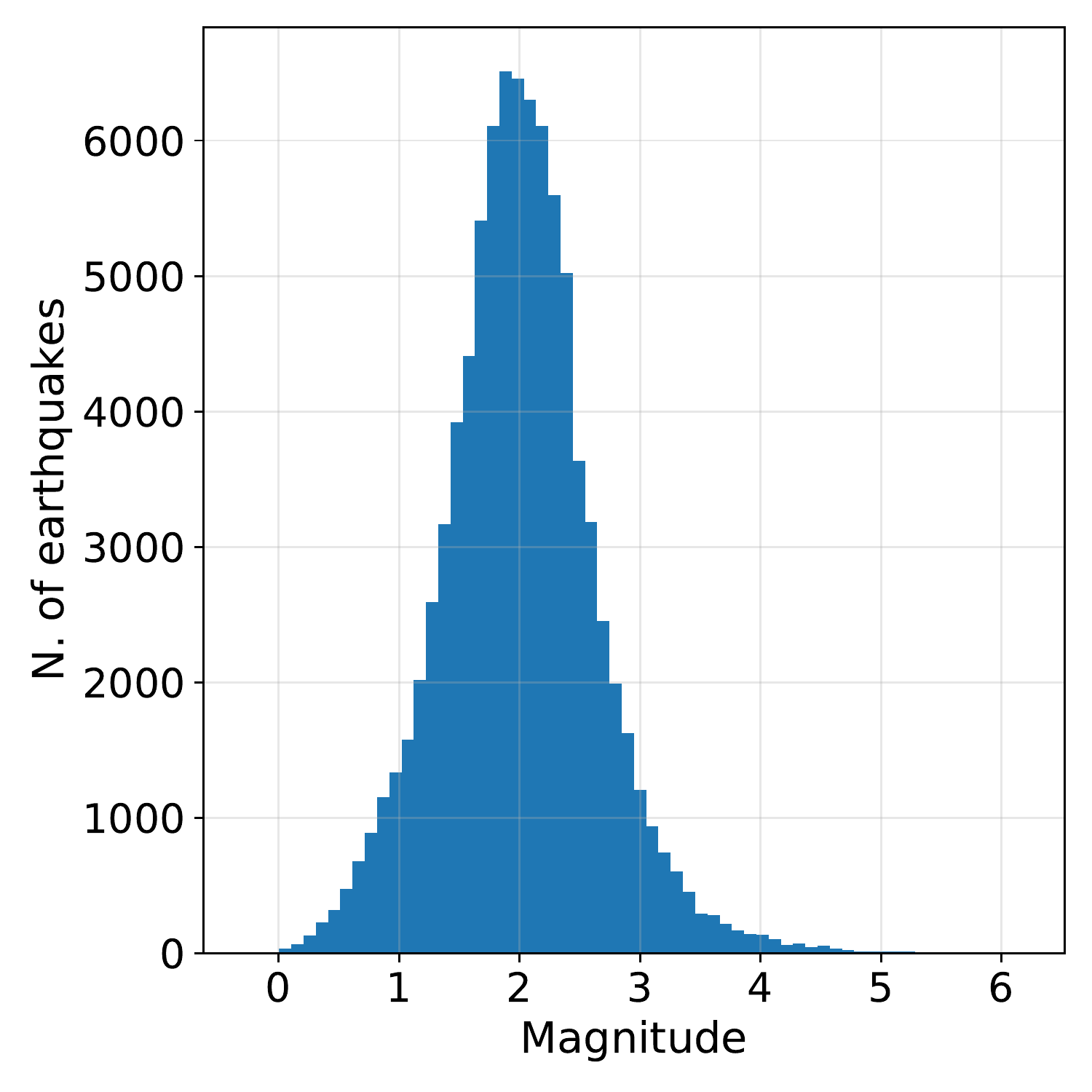}
       \includegraphics[width=0.49\columnwidth]{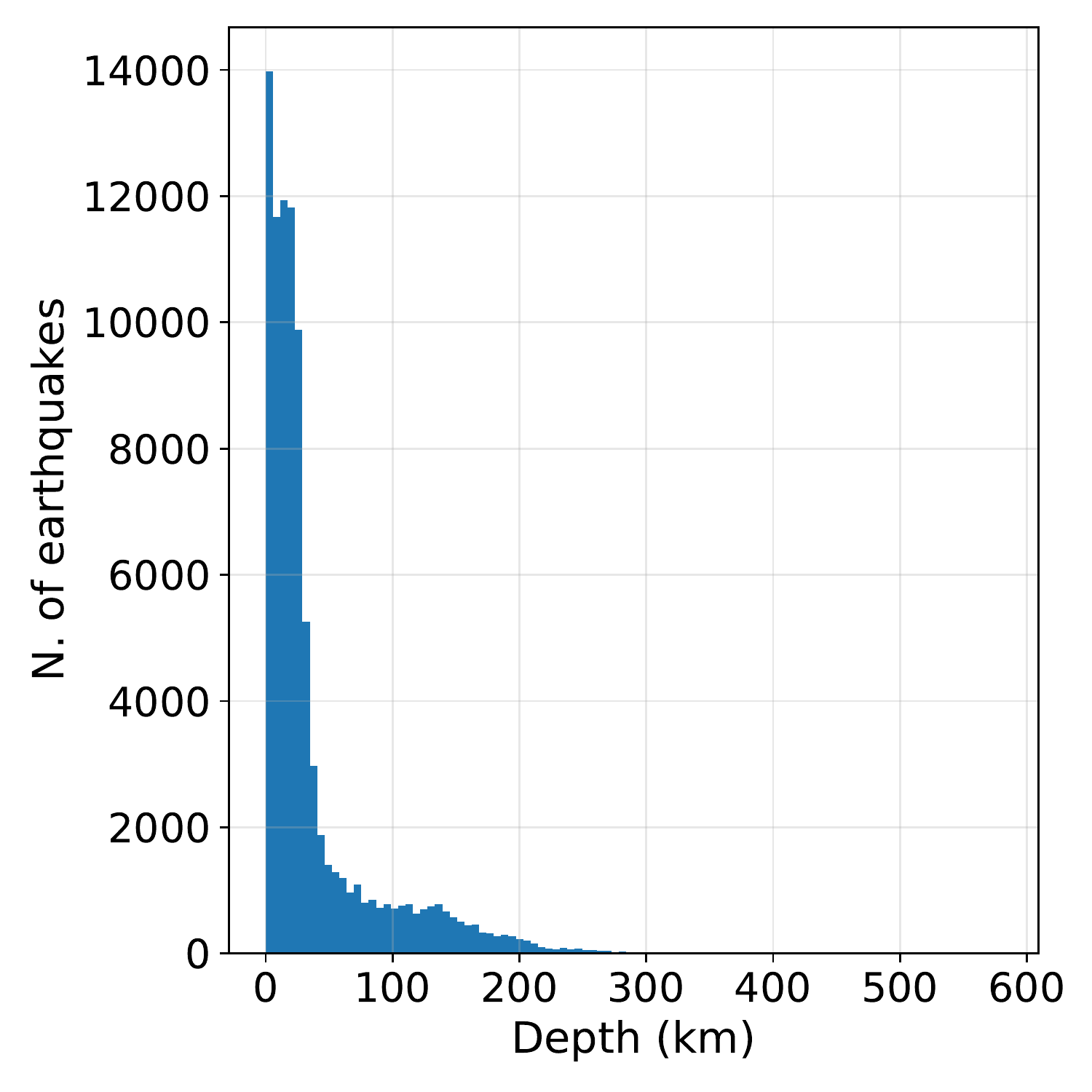}\hfill
\caption{(Left) Magnitude Distribution. (Right) Depth Distribution.}\vskip-.45cm
\label{fig:EQcraul}
\end{figure}

\begin{figure}[t]
\centering
        \includegraphics[width=\columnwidth]{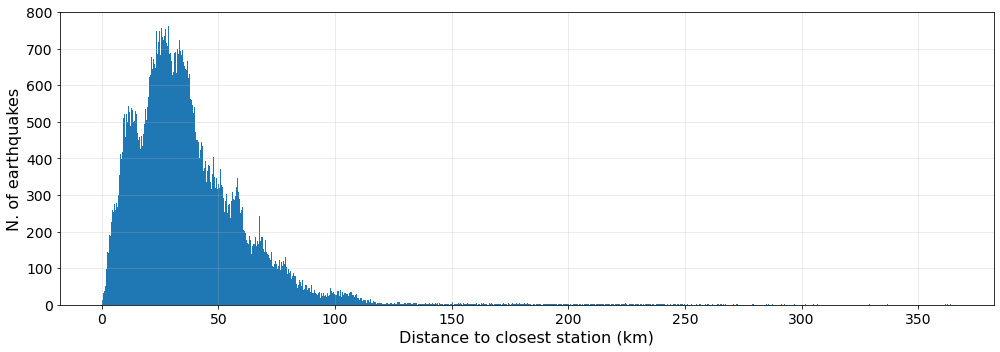}
         \includegraphics[width=\columnwidth]{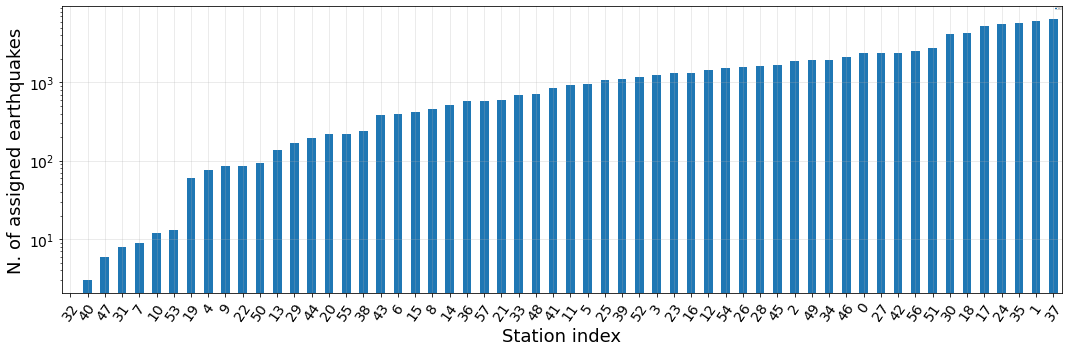}
\caption{(Top) Distribution of the epicenter distance from the closest station. (Bottom) Number of assigned earthquakes per station.}\vskip-.45cm
\label{fig:EQcraul1}
\end{figure}

\smallskip
\noindent\textbf{Dataset.} We extracted $90,000$ initial recordings between 2016 and 2020 over $58$ seismic stations; Fig.~\ref{fig:EQdistrib}. The signal is the weak motion measured along the vertical axis at $100$Hz. We kept only those recordings for which the magnitude was between one and three [cf. Fig.~\ref{fig:EQcraul} (left)] and the depth smaller than $200$ km [cf. Fig.~\ref{fig:EQcraul} (right)]. This resulted into a more uniform distribution with $87,000$ recordings. We further discarded those datapoints for which one of the stations was inactive and retained those earthquakes for which the epicentre was within $75$km from the closest station, leaving to approximately $70,000$ datapoints; Fig.~\ref{fig:EQcraul1} (top). The dropped recordings are mostly in the ocean.

We further analyzed the number of assigned earthquakes per station Fig.~\ref{fig:EQcraul1} (bottom). There was one station with no earthquakes assigned; several stations with less than $100$; and $27$ stations with more than $1,000$. This distribution leads to an unbalanced dataset, which was prioritising the majority classes even by using conventional learning approaches for unbalanced datasets. To achieve a more balanced dataset, we discarded those stations with less than $150$ earthquakes and undersampled randomly the recorded seismic waves in those stations containing more than $1,000$ recordings. This led to the final dataset comprising $4,633$ recordings assigned to $45$ stations (out of the 58 available), while the graph remains defined over the $N = 58$ stations.

\smallskip
\noindent\textbf{Graph construction.} We built a geometric graph following the great-circle distance strategy with an edge if the distance between two stations is smaller than $170,3$ km. We further weighted the graph by setting the edge weights as $A_{ij} = e^{-d(i,j)}/\bar{d}$, where $d(i,j)$ is the great-circle distance between stations $i$ and $j$ and $\bar{d}$ is the average distance. {Refer to Fig.~5 in the main document.}

\begin{figure}[!t]
\centering
       \includegraphics[width=\columnwidth]{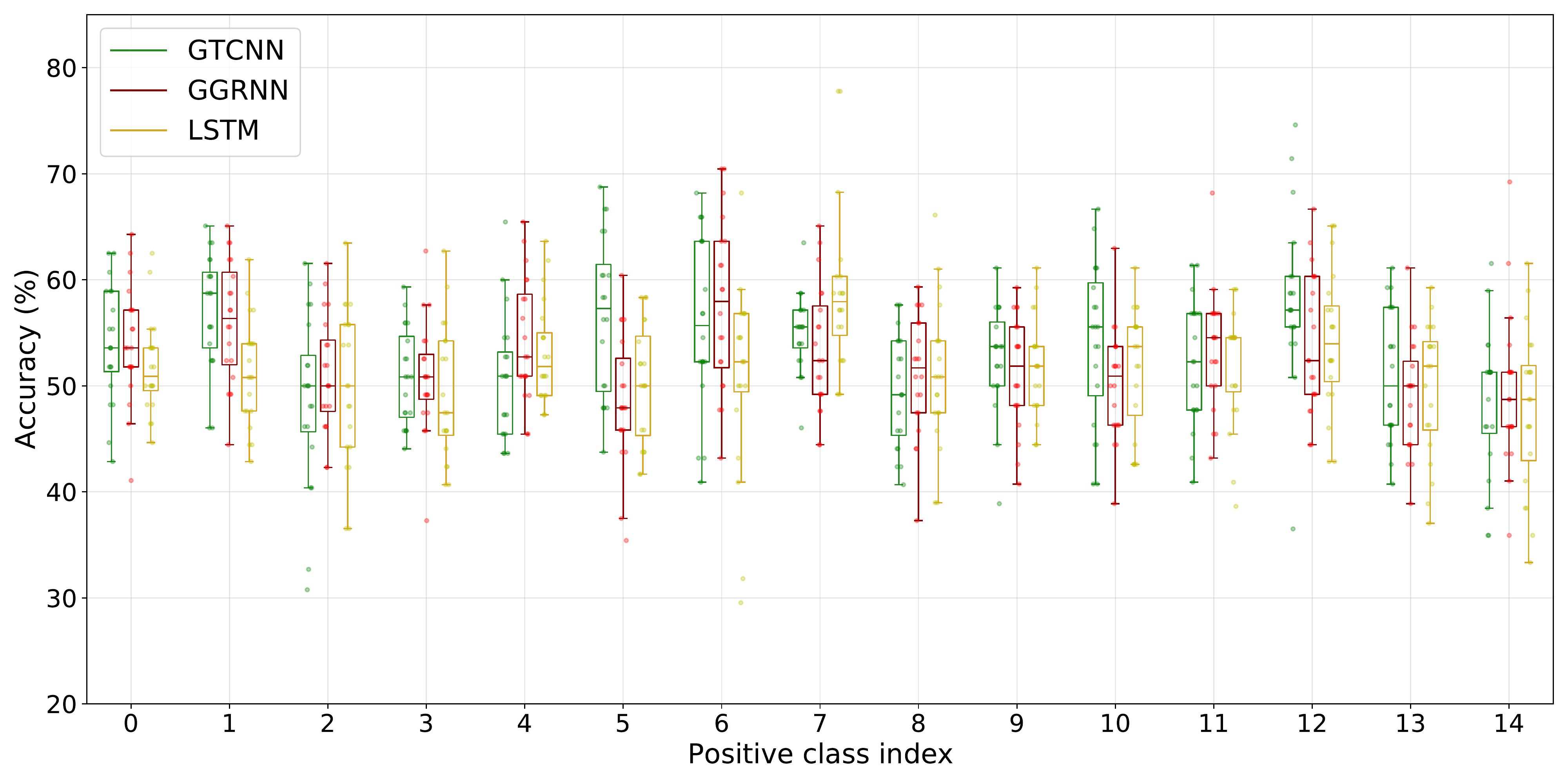}
       \includegraphics[width=\columnwidth]{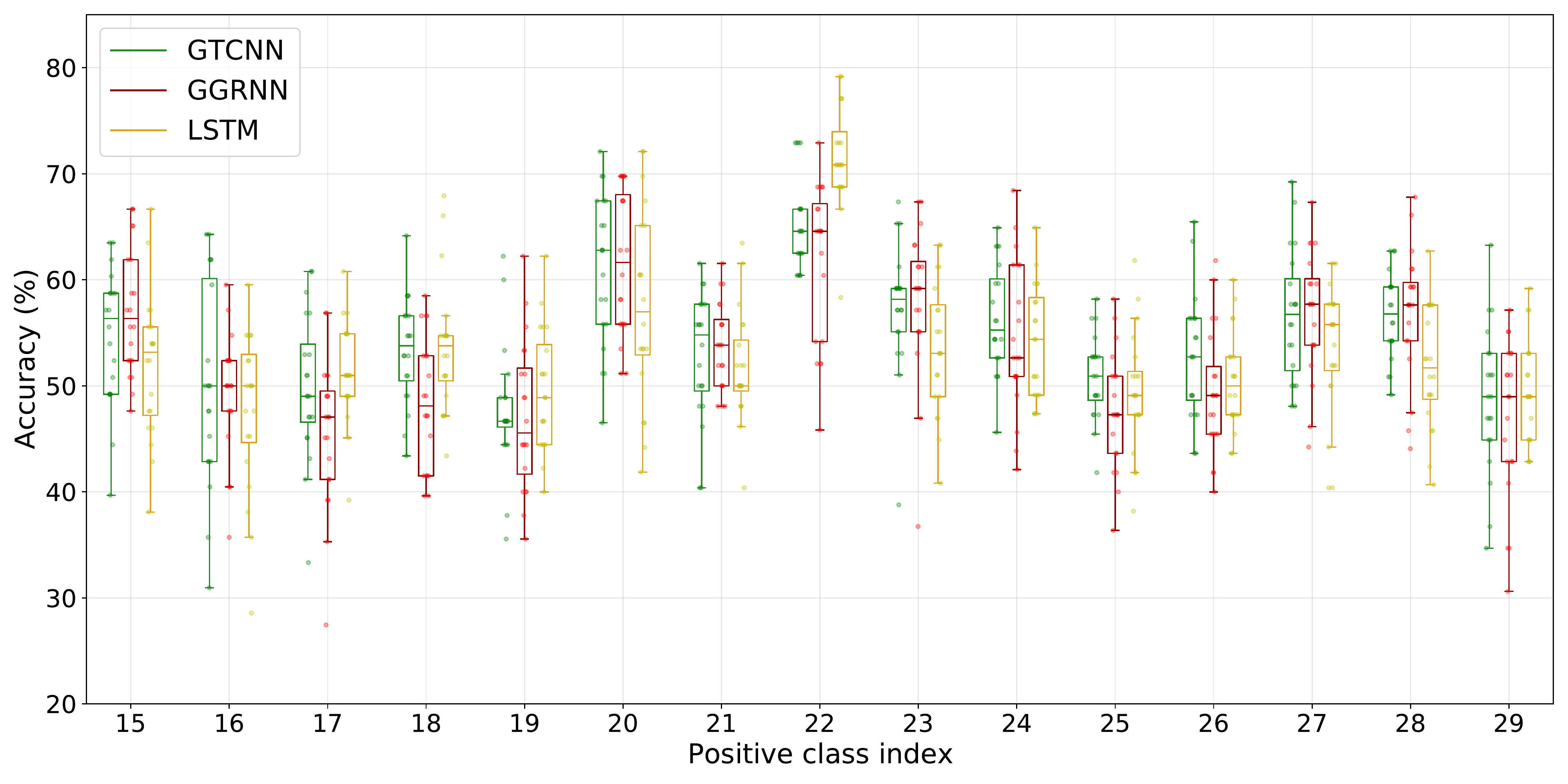}
       \includegraphics[width=\columnwidth]{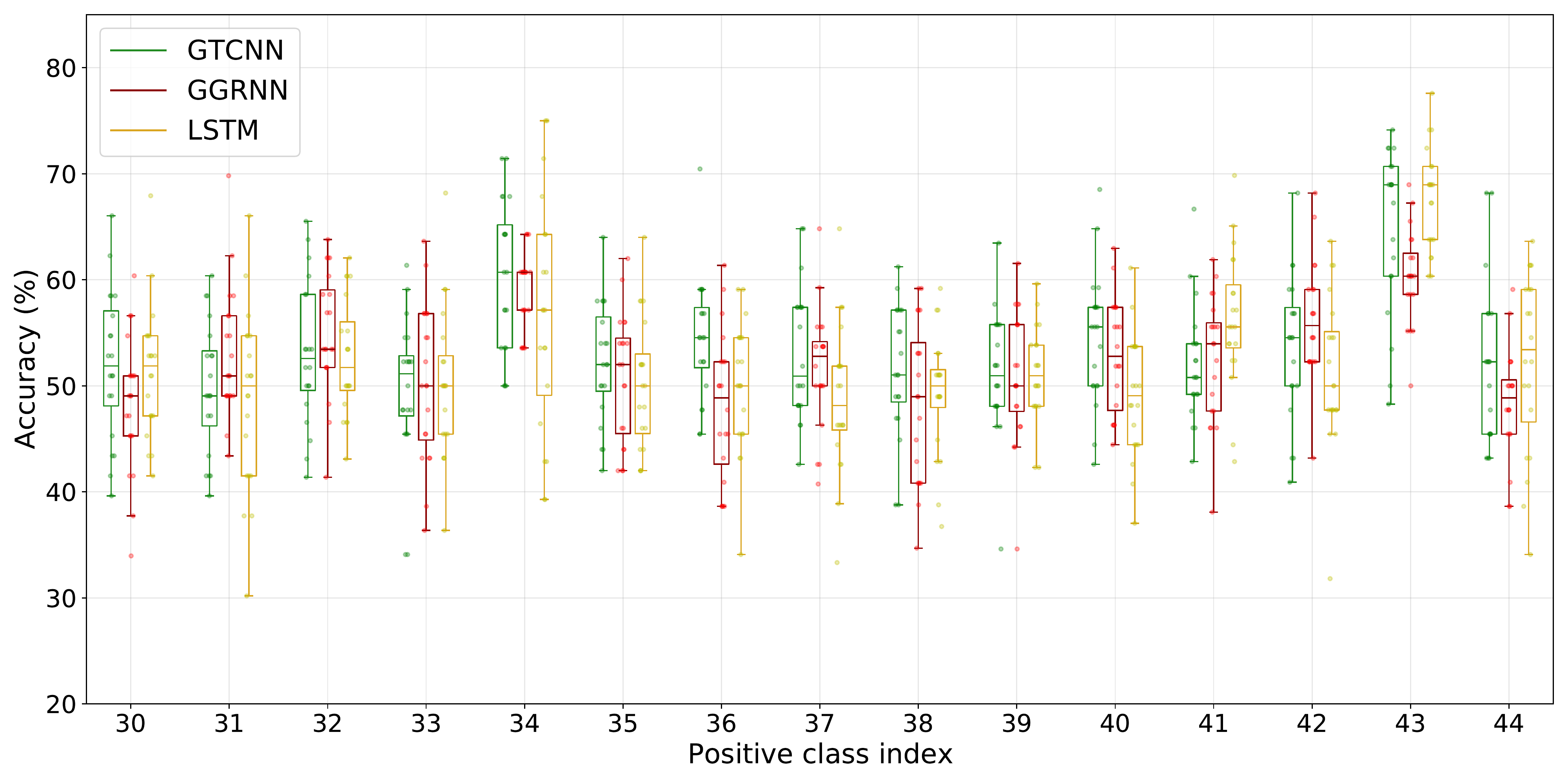}
\caption{Accuracies of the proposed GTCNN and alternatives GGRNN and LSTM for the different positive classes (stations) in the one-vs-all classification.}\vskip-.6cm
\label{fig:GTCNNOne-vs-all}
\end{figure}

\smallskip
\noindent\textbf{One-vs-all.} Fig.~\ref{fig:GTCNNOne-vs-all} shows the boxplot distributions for the different classes. We can see the GTCNN can reach a median accuracy of about $60\%$ in $11$ cases and also hitting up to $70\%$ in a few of them. Nevertheless, the scarcity of the data, the uneven spatial distribution of the earthquakes, and difficulty of working with recordings before the actual earthquake make it difficult for all methods to achieve far superior accuracies. Further research will be based on this aspect and assessing the role of the graph for this setting.

\end{document}